\documentclass[10pt,twocolumn,letterpaper]{article}

\usepackage[main]{cvpr}      
\definecolor{cvprblue}{rgb}{0.21,0.49,0.74}
\usepackage[accsupp]{axessibility}
\usepackage[pagebackref,breaklinks,colorlinks,allcolors=cvprblue]{hyperref}
\usepackage[ruled,vlined,linesnumbered]{algorithm2e}
\usepackage{amsmath}
\usepackage{multirow}
\usepackage{array}
\usepackage{tocloft}
\def\paperID{*} 
\def\confName{CVPR}
\def\confYear{2026}

\title{Adaptive Video Distillation: Mitigating Oversaturation and Temporal Collapse in Few-Step Generation}

\author{
Yuyang You$^{1,}$\thanks{Work done during an internship at Kuaishou.} \quad
Yongzhi Li$^{2,}$\thanks{Project Leader, $^\ddagger$Corresponding authors.} \quad
Jiahui Li$^{2}$ \quad
Yadong Mu$^{1,\ddagger}$ \quad
Quan Chen$^{2,\ddagger}$ \quad
Peng Jiang$^{2}$ \\
$^{1}$Peking University \quad $^{2}$Kuaishou Technology \\
{\tt\small yyyou25@stu.pku.edu.cn} ~ {\tt\small myd@pku.edu.cn} \\
{\tt\small \{liyongzhi03,lijiahui11,chenquan06,jiangpeng\}@kuaishou.com}
}

\begin{document}
\maketitle
\addtocontents{toc}{\protect\setcounter{tocdepth}{-10}}

\vspace{-0.1in}
\begin{abstract}
Video generation has recently emerged as a central task in the field of generative AI. However, the substantial computational cost inherent in video synthesis makes model distillation a critical technique for efficient deployment. Despite its significance, there is a scarcity of methods specifically designed for video diffusion models. Prevailing approaches often directly adapt image distillation techniques, which frequently lead to artifacts such as oversaturation, temporal inconsistency, and mode collapse. To address these challenges, we propose a novel distillation framework tailored specifically for video diffusion models. Its core innovations include: (1) an \textbf{adaptive regression loss} that dynamically adjusts spatial supervision weights to prevent artifacts arising from excessive distribution shifts; (2) a \textbf{temporal regularization loss} to counteract temporal collapse, promoting smooth and physically plausible sampling trajectories; and (3) an \textbf{inference-time frame interpolation} strategy that reduces sampling overhead while preserving perceptual quality. Extensive experiments and ablation studies on the VBench and VBench2 benchmarks demonstrate that our method achieves stable few-step video synthesis, significantly enhancing perceptual fidelity and motion realism. It consistently outperforms existing distillation baselines across multiple metrics. The source code is publicly available at: \url{https://github.com/yuyangyou/Adaptive-Video-Distillation}
\end{abstract}
\vspace{-0.1in}    
\vspace{-0.15in}
\section{Introduction}

\label{sec:intro}
Diffusion models have emerged as the cornerstone of modern generative AI, achieving remarkable progress in both image and video synthesis~\cite{ho2020denoising,song2020score,rombach2022high,blattmann2023stable,ho2022imagen,karras2022elucidating,poole2022dreamfusion,esser2023structure,bao2024vidu}. Compared with traditional generative paradigms such as GANs~\cite{goodfellow2014generative}, diffusion models demonstrate superior performance in fidelity, diversity, training stability, and scalability. However, their inference latency remains a major bottleneck for real-world deployment. 
This is especially significant in video generation, where existing models often process tens of thousands of tokens simultaneously in each denoising step and repeat this process for numerous iterations, incurring enormous computational overhead.
Although training-free acceleration strategies based on specialized samplers~\cite{lu2022dpm,zhang2022fast,song2020denoising} can improve efficiency, they still require dozens of sampling steps to produce satisfactory results due to the inherent discretization errors of numerical solvers. In contrast, training-based distillation methods enable few-step or even single-step generation while maintaining high-quality outputs.

\begin{figure}[t]
    \centering
    \includegraphics[width=1\linewidth, height=0.3\textheight, keepaspectratio]{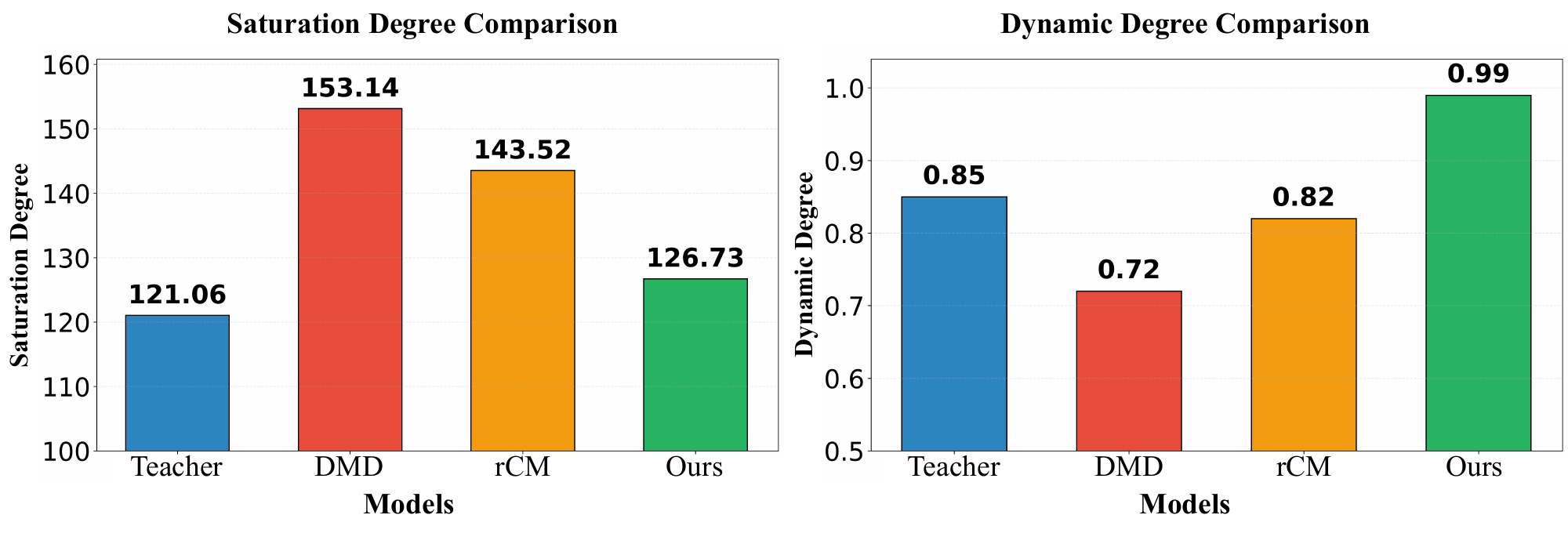}
    \caption{\small Baselines like DMD and rCM show severe color oversaturation (left) and reduced motion indicative of temporal collapse (right). Our method achieves appropriate saturation and enhances motion dynamics beyond the teacher model.}
    \label{fig:fig1}
    \vspace{-0.25in}
\end{figure}
Among existing approaches, Distribution Matching Distillation (DMD)~\cite{yin2024one,yin2024improved} has achieved great success in industrial applications, owing to its strong capability in preserving fine-grained image details. Nevertheless, the naive application of DMD usually suffers from oversaturation and mode collapse~\cite{lu2024simplifying,yin2024one,yin2024improved,zheng2025large}. These issues become significantly more severe when extended to video generation (Figure~\ref{fig:fig1}). In autoregressive video diffusion models distilled via DMD, temporal error accumulation exacerbates the saturation issue, leading to noticeable quality degradation in long video generation~\cite{cui2025self,teng2025magi,liu2026streamingautoregressivevideogeneration}. Meanwhile, mode collapse extends into the temporal dimension, resulting in videos with limited or even static motion.

To address these challenges, we introduce a novel distillation framework built upon two key components: an Adaptive Regression Loss and a Temporal Regularization Loss. The Adaptive Regression Loss empowers the student model to more effectively align with the teacher's distribution, while simultaneously and selectively integrating supervision from the real data manifold. This dual-guidance strategy steers the student toward a more optimal target distribution, which effectively resolves the excessive oversaturation problem. Concurrently, the Temporal Regularization Loss is specifically designed to counteract the degradation of motion dynamics during distillation. It ensures that temporal coherence and motion fidelity are preserved, thereby preventing the temporal mode collapse that leads to static or motion-limited video generation.

Crucially, the Adaptive Regression Loss is designed with a dual purpose. Beyond distillation, it intrinsically supports simultaneous supervised fine-tuning, allowing the model to be directly migrated to specialized video distributions, such as anime, advertisements, or visual effects during the distillation process. This capability grants our method superior adaptability for practical applications, overcoming the inefficiency and performance degradation of traditional methods, which require a sequential process of fine-tuning the teacher model and then performing standard distillation.

To further enhance the practicality, we analyze the diffusion process and identify that high-noise steps primarily generate high-level semantics, exhibiting minimal temporal feature variance. Based on this insight, we design a novel decoupled temporal interpolation module. This module strategically performs inference at a lower frame rate during high-noise steps and subsequently interpolates back to the original frame rate in low-noise steps. This optimization significantly reduces the computational load for inference with negligible impact on generation quality.

We conduct extensive experiments and ablation studies on the open-source video diffusion model Wan 2.1. The proposed framework enables the efficient training of few-step student models capable of generating realistic videos with fine spatial details, natural colors, and coherent motion dynamics. On the VBench benchmark, our method consistently surpasses state-of-the-art baselines across multiple evaluation dimensions, showcasing both strong quantitative performance and substantial potential for practical deployment.

\vspace{-0.1in}

\section{Related Work}
\label{sec:formatting}
\vspace{-0.05in}
\subsection{Video Generation and Diffusion Models}
Early video generation primarily relied on Generative Adversarial Networks (GANs)~\cite{clark2019efficient,tulyakov2018mocogan,vondrick2016generating}, 
which introduced 3D spatiotemporal convolutions, content-motion disentanglement, and hierarchical discriminators. 
More recently, Diffusion Models (DMs)~\cite{ho2020denoising,song2020denoising,nichol2021improved,ho2022video} have emerged as the state-of-the-art paradigm, 
outperforming GANs in training stability, sample diversity, and perceptual fidelity through iterative denoising.
A pivotal advancement is Latent Diffusion Models (LDMs)~\cite{rombach2022high}, 
which perform denoising in VAE-projected latent space rather than pixel space, 
drastically reducing computational overhead while preserving fine-grained spatiotemporal details. 
Building on this, Flow Matching and Rectified Flow~\cite{lipman2022flow,liu2023rectified} formulate diffusion as continuous ODEs 
with straighter trajectories between noise and data, theoretically enabling fewer sampling steps.
State-of-the-art open-source video generation models~\cite{villegas2023phenaki,chen2024videocrafter2,yan2024moviegen,blattmann2023stable,brooks2024video,wan2025wan,seawead2025seaweed,bao2024vidu} are predominantly built upon flow or latent-space-based modeling approaches.
In parallel, Autoregressive (AR) approaches~\cite{yan2021videogpt,yu2023magvitmaskedgenerativevideo,blattmann2024stable,yin2025slow,huang2025self,cui2025self} 
leverage Transformers with causal attention to sequentially generate frames or tokens, supporting variable-length synthesis. 
However, they suffer from error accumulation, leading to temporal inconsistency and motion decay in longer sequences.
While significant strides have been made, a critical bottleneck persists: the substantial computational demand of leading video generation models. Specifically, diffusion models built upon Transformers~\cite{peebles2023scalable} and classifier-free guidance~\cite{ho2022classifier} necessitate tens to hundreds of denoising iterations to synthesize even short video segments (5-10 seconds). This computational burden renders them impractical for real-time interactive applications. To address this limitation, our research pivots towards distillation methods, seeking to drastically reduce inference time while maintaining the fidelity of the generated content.

\begin{figure*}[t]
    \centering
    \includegraphics[width=1.0\textwidth]{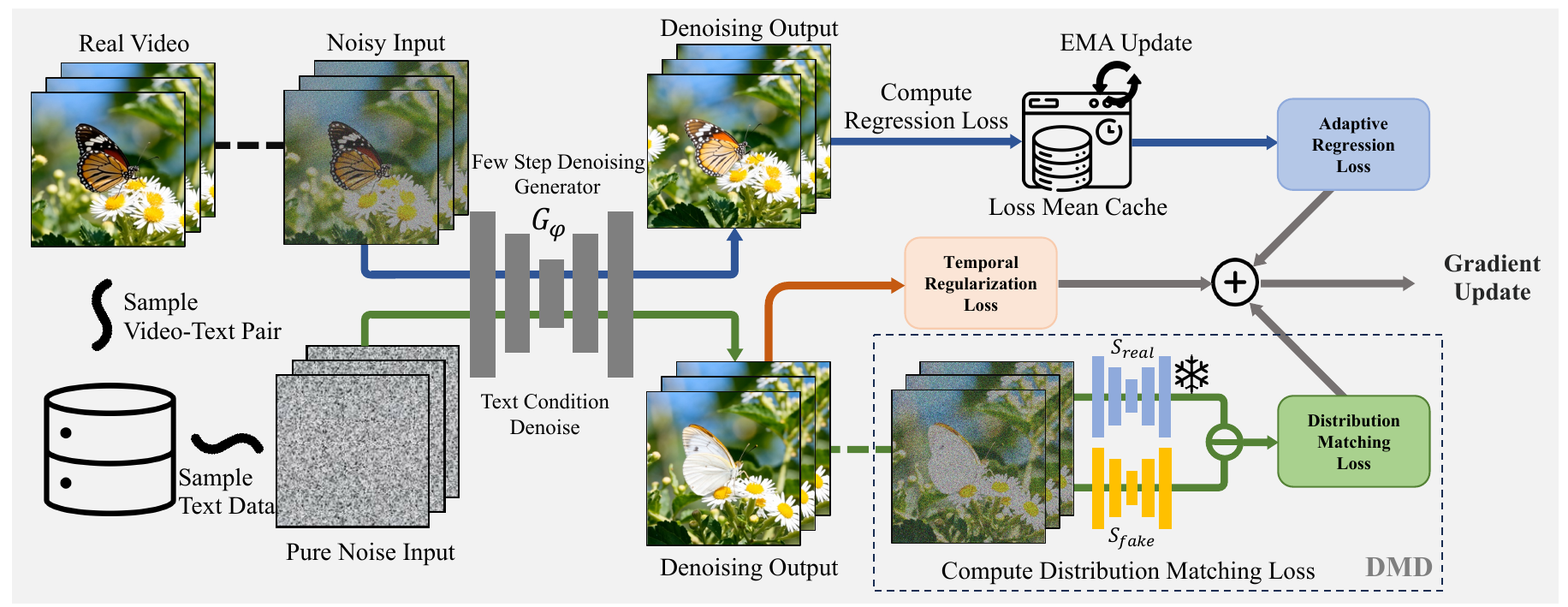}
    \caption{\small Our method distills a pre-trained teacher model, denoted as $s_{\text{data}}$, into a few-step video generator $G_\phi$. The training procedure consists of the following steps:(1) A batch of real video-text pairs is sampled from the dataset. After applying noise perturbations to the videos, the student model performs denoising reconstruction. A regression loss is computed between the reconstructed video and the ground-truth video. Subsequently, this loss is adaptively weighted using our Loss Mean Cache to produce the final adaptive regression loss (see Sec.~\ref{subsec:temp} for details). (2) Text conditions are sampled from the dataset to guide the student model in generating a video from pure noise. The denoised output from this process is used to compute a temporal regularization loss (Eq.~\ref{eq:temp}) and a distribution matching loss (Eq.~\ref{eq:dmd}).(3) Finally, the generator $G_\phi$ is updated via gradient descent using the combined losses. The $s_{\text{gen}, \xi}$ in DMD are updated separately, following the methodology of DMD2 (this particular update step is not depicted in the figure for clarity).
        }
    \label{fig:full_architecture}
    \vspace{-0.2in}
\end{figure*}
\subsection{Diffusion Distillation}

Diffusion model distillation has evolved rapidly, giving rise to multiple complementary paradigms that progressively improve generation efficiency and fidelity.
Early studies adopt knowledge distillation~\cite{hinton2015distilling}, which directly supervises the prediction of noise or score functions to transfer the denoising behavior of a pretrained teacher model to a lightweight student.
Building upon this idea, progressive distillation~\cite{salimans2022progressive,meng2023distillation,lin2024sdxl} introduces a recursive framework that halves the sampling steps at each stage, allowing the student to approximate the teacher’s denoising trajectory from coarse to fine.
In parallel, adversarial distillation~\cite{sauer2024adversarial,sauer2024fast,xu2024ufogen,lin2024sdxl,kang2024distilling,lin2025diffusion,luo2024you,luo2025adding} explicitly incorporates adversarial objectives to enhance perceptual fidelity and texture sharpness.More recently, consistency distillation~\cite{song2023consistency,luo2023latent,kim2023consistency,lee2024truncated,heek2024multistep,ren2024hyper,lv2025dcm,wang2024phased,lu2024simplifying,zheng2025large} reformulates diffusion training as a path-matching problem between noisy and clean samples, enabling continuous-time modeling without relying on discrete solvers.
This paradigm promotes better mode coverage and can generate videos with high diversity, though often at the expense of fine-detail fidelity.
Meanwhile, score distillation~\cite{poole2022dreamfusion,wang2023prolificdreamer,zhou2024score,luo2024one,luo2023diff,yin2024one,yin2024improved,salimans2024multistep,xie2024distillation,luo2025learning,xu2025one,fan2025phased} has emerged as a unifying framework for distribution alignment, optimizing the student model to match the teacher’s score field using statistical distance measures such as reverse KL divergence.
A representative example is Distribution Matching Distillation (DMD), which achieves strong fine-grained detail synthesis but suffers from mode collapse and over-saturation issues.
Several works attempt to mitigate these problems by combining DMD with adversarial objectives~\cite{yin2024improved,lu2025adversarial} or jointly training it with consistency-based losses~\cite{zheng2025large} in image generation.
In contrast, our method builds upon distribution matching distillation, introducing an efficient and stable supervised regularization strategy that effectively addresses these issues for video generation.

\vspace{-0.1in}
\section{Method}
\vspace{-0.1in}
In this section, we first analyze the limitations of distribution-matching distillation in video generation (Sec.~\ref{subsec:3.1}). We present a novel training procedure to tackle these challenges, as shown in Figure~\ref{fig:full_architecture}. Its novelty lies in two fundamental components: an adaptive regression loss (Sec.~\ref{subsec:adpt}) that enhances the learning stability and accuracy, and a temporal regularization mechanism (Sec.~\ref{subsec:temp}) that enforces temporal consistency across frames. In addition, we present an inference-time frame interpolation acceleration strategy (Sec.~\ref{subsec:3.4}) to enable efficient generation of high-quality videos.

\subsection{Analysis of DMD Methods}
\vspace{-0.05in}

\label{subsec:3.1}
In a standard diffusion process, for samples drawn from the data distribution $p(x_0)$, the forward diffusion process is defined as
\vspace{-0.02in}
\begin{equation}
x_t = \alpha_t x_0 + \sigma_t \epsilon, \quad \epsilon \sim \mathcal{N}(0, I),
\label{eq:forward}
\end{equation}
\vspace{-0.02in}
where $x_t$ denotes the noisy sample, $\alpha_t$ and $\sigma_t$ are scalar coefficients that determine the noise schedule and the signal-to-noise ratio at timestep $t$. The diffusion model learns a reverse denoising process by minimizing the objective
\vspace{-0.02in}
\begin{equation}
\mathcal{L}(\theta) = \mathbb{E}_{t, x_0, \epsilon} \left[ \| \epsilon_\theta(x_t, t) - \epsilon \|_2^2 \right],
\label{eq:mse_loss}
\end{equation}
\vspace{-0.02in}
where \(\epsilon_\theta(x_t, t)\) is a neural network parameterized by $\theta$ that predicts the noise at timestep $t$. Flow-matching models instead predict the instantaneous velocity field, but under certain conditions, the two formulations are equivalent. To recover a clean sample \(x_0\) from a noisy initialization $x_T$ drawn from a Gaussian prior, the reverse process requires not only accurate noise prediction but also numerical solvers to simulate the denoising trajectory.
Score distillation leverages pretrained diffusion models as differentiable priors to transfer generative capabilities into compact few-step generators. The diffusion model parameterized by \(\theta\) implicitly defines the data distribution through its score function
\vspace{-0.1in}
\begin{equation}
s_\theta(x_t, t) = \nabla_{x_t} \log p(x_t) = -\frac{1}{\sigma_t} \, \epsilon_\theta(x_t, t),
\label{eq:score}
\end{equation}
\vspace{-0.02in}
where \(\epsilon_\theta(x_t, t)\) denotes the noise prediction at timestep \(t\). This score function provides gradients pointing toward high-density regions of the data distribution, serving as a foundation for distillation objectives.
Formally, score-based distillation minimizes a divergence \(D(p_{\text{gen}, t} \| p_{\text{data}, t})\) between the student generator’s marginal \(p_{\text{gen}, t}\) and the teacher diffusion model’s distribution \(p_{\text{data}, t}\). Taking DMD as an example, it's gradient objective can be expressed as
\begin{equation}
\begin{split}
\nabla_\phi \mathcal{L}_{\text{DMD}} 
&\triangleq 
\mathbb{E}_t \bigl[ \nabla_\phi \, \mathrm{KL}\bigl(p_{\text{gen}, t} \| p_{\text{data}, t}\bigr) \bigr] \\
&\approx
- \mathbb{E}_t 
\biggl[ \int
\biggl( s_{\text{real}}\bigl(\Psi(G_\phi(\epsilon), t), t\bigr) \\
&\quad - s_{\text{fake}}\bigl(\Psi(G_\phi(\epsilon), t), t\bigr)
\biggr)
\frac{\partial G_\phi(\epsilon)}{\partial \phi}
\, d\epsilon \biggr],
\end{split}
\label{eq:dmd}
\end{equation}
where \(s_{\text{real}}\) and \(s_{\text{fake}}\) denote the score functions of the teacher model and an online model trained on the generator’s output distribution, respectively. And \(\Psi(\cdot)\) represents the forward diffusion process.

\begin{figure}[t]
    \centering
    \includegraphics[width=0.9\linewidth]{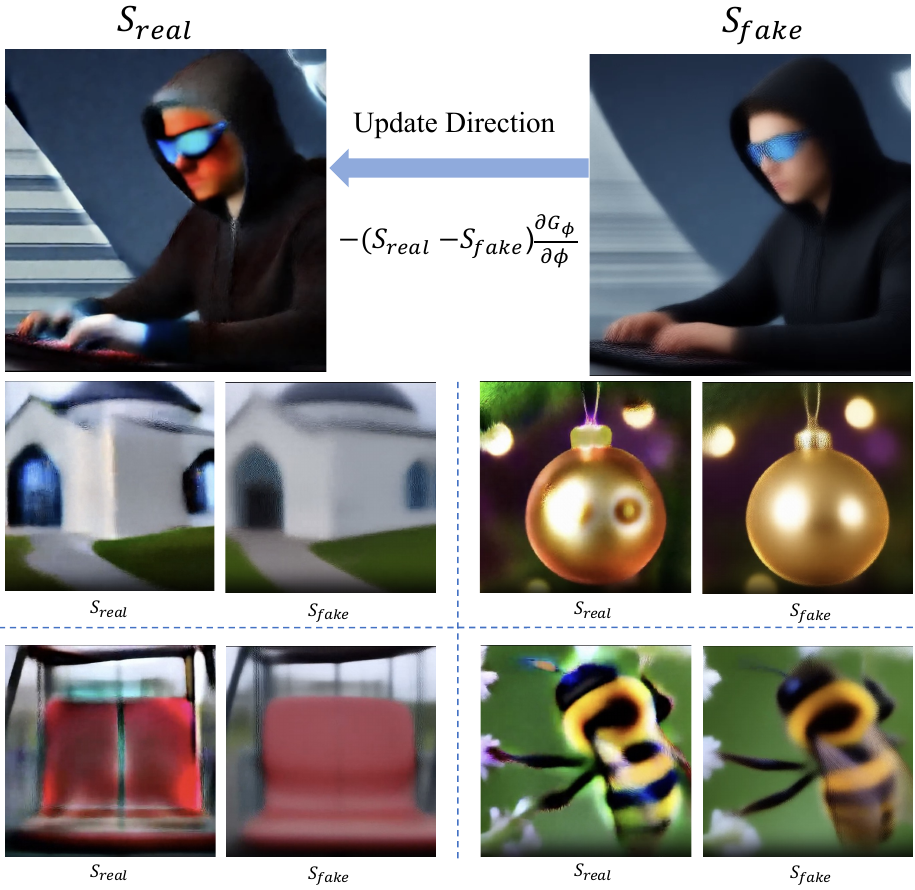}
    \caption{
This figure explains the origin of oversaturation in distribution-matching distillation. At a given denoising timestep, $s_{\text{real}}$ denotes the teacher’s ground-truth clean-sample distribution and $s_{\text{fake}}$ the student’s estimate obtained via an online multi-step model; the teacher’s overemphasis on fine-grained detail biases the student toward an oversaturated, suboptimal distribution, degrading perceptual video quality.
}
\vspace{-0.1in}
\label{fig:score}
\end{figure}
\vspace{-0.1in}
\paragraph{Oversaturation Problem.} The teacher score provides stable gradients that guide the student generator \(G_\phi\) toward matching the target data distribution. However, in practice, training the student model solely with the distribution matching loss often leads to oversaturated visual outputs~\cite{yin2024improved}. As illustrated in Figure~\ref{fig:score}, this occurs because the teacher model, while encouraging the student to capture finer details, tends to overemphasize local information, steering the student toward a suboptimal distribution characterized by excessive color saturation. This issue becomes particularly severe in autoregressive generation, where oversaturation can cause error accumulation across frames~\cite{huang2025self,cui2025self}. To mitigate this, we introduce a direct supervision mechanism for the student model in Sec.~\ref{subsec:adpt}, effectively correcting this bias and alleviating the oversaturation problem in generated videos.
\paragraph{Temporal Mode Collapse.}Moreover, distribution matching losses are known to induce mode collapse in image generation~\cite{yin2024one,zheng2025large,lu2025adversarial}, reducing diversity across samples. In video generation, this problem is further amplified by the temporal dimension, manifesting as diminished motion amplitude or even static outputs. Since insufficient motion often degrades perceived video quality more severely than reduced spatial diversity, we introduce a temporal regularization constraint in Sec.~\ref{subsec:temp} to explicitly address temporal collapse and enhance motion dynamics in the generated sequences.


\begin{figure}
    \centering
    \includegraphics[width=1\linewidth]{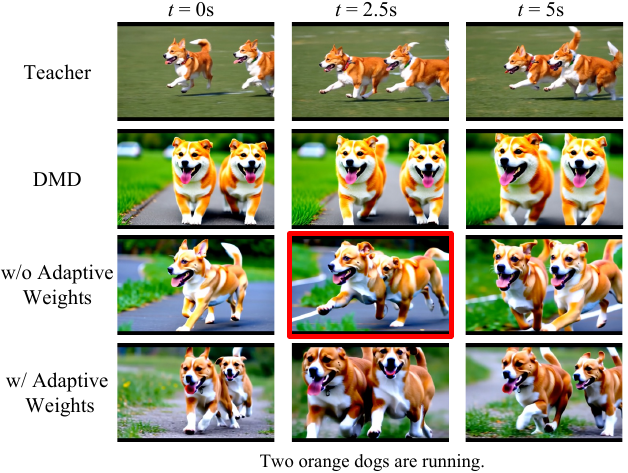}
    \vspace{-0.2in}
    \caption{A naive regression loss (Row 3) causes object fusion artifacts (t=2.5s) absent in the teacher (Row 1) and baseline (Row 2). Our adaptive loss (Row 4) resolves this artifact, improving generation quality.}
    \label{fig:two dogs}
    
\end{figure}

\subsection{Adaptive Regression Loss}
\label{subsec:adpt}
Although the teacher model provides strong gradient guidance in high-quality regions, the distribution matching distillation effectively transfers the teacher’s ability to synthesize fine-grained details. However, this detail-oriented optimization often leads to oversaturation, especially in autoregressive generation, where such artifacts accumulate over time.

To alleviate this issue, we introduce a regression loss that directly supervises the student model, injecting real data information to jointly guide the gradient updates alongside the distillation loss. Nevertheless, when the student model simultaneously fits the teacher distribution and is influenced by real samples that deviate significantly from it, the optimization may converge to a suboptimal intermediate distribution. As illustrated in Figure~\ref{fig:two dogs}, this can cause visual artifacts such as tearing, unnatural object blending, duplication, or disappearance.

To address this problem, we propose an adaptive regression loss, which flexibly regularizes the student output toward real data samples while suppressing the influence of inconsistent data points by assigning them lower weights. The loss is defined as
\begin{equation}
\mathcal{L} = w_{t,s} \, \| \epsilon_\theta(x_t, t) - \epsilon \|_2^2,
\label{eq:regloss}
\end{equation}
where \(w_{t,s}\) is a learnable, timestep-dependent adaptive weight. For real video data sampled at different locations, the deviation between the generated and true distributions varies significantly. Points with larger deviations tend to cause overly aggressive gradient updates, resulting in tearing artifacts or suboptimal frames. To alleviate this, we assign lower weights to such points.
We maintain a dynamic, learnable cache to estimate the expected loss value at each timestep \(t\) of the student’s few-step inference schedule. This cache is updated using an exponential moving average (EMA):
\begin{equation}
\bar{\mathcal{L}}_{t,s} = \alpha \, \mathcal{L}_{t,s-1} + (1 - \alpha) \, \mathcal{L}_{s},
\label{eq:ema}
\end{equation}
where \(\alpha\) denotes the EMA coefficient, \(t\) indexes the diffusion timestep, and \(s\) represents the current training iteration. Based on this cache, the adaptive gradient weight for the current sample is computed as
\begin{equation}
\omega_{t,s} = 1 - \sigma \big( k \cdot (\mathcal{L}_s - \bar{\mathcal{L}}_{t,s-1}) \big),
\quad
\sigma(x) = \frac{1}{1 + e^{-x}},
\label{eq:weight}
\end{equation}
where \(k\) is a scaling factor and \(\sigma(\cdot)\) is the Sigmoid function.
This adaptive weighting mechanism enables the student model to smoothly align with the real data distribution while suppressing overexposure. Moreover, it improves both temporal and spatial diversity, effectively alleviating mode collapse that often arises during few-step distillation
\subsection{Temporal Regularization}
\label{subsec:temp}
\vspace{-0.05in}

As discussed in Sec.~\ref{subsec:3.1}, compared to image generation, the impact of mode collapse on the additional temporal dimension in video generation is non-negligible. This phenomenon often manifests as limited motion dynamics, sometimes even near-static sequences or reduced variations in object appearances (See appendix Sec.~\ref{sec:Visualize} for visual examples). In certain application scenarios, such degradation in temporal coherence can be far more detrimental to video quality than the loss of spatial diversity typically caused by mode collapse.
The adaptive regression loss introduced in the previous section partially alleviates spatial mode collapse by incorporating real data supervision, enabling the student to explore unseen modes. However, its supervision over the temporal dimension remains weak. Compared with reduced object diversity, the lack of motion typically has a far more detrimental effect on perceived video quality.
To address this, we introduce a motion regularization loss that explicitly encourages temporal variation in the generated sequences:
\begin{equation}
\mathcal{L}_\text{temp} = - \log \big( \mathbb{E}_{x \sim p_\theta} [ \mathrm{Var}(x) ] + \epsilon \big),
\label{eq:temp}
\end{equation}
where \(x\) denotes samples generated by the student model, and the variance \(\mathrm{Var}(x)\) is computed along the temporal dimension. The constant \(\epsilon\) ensures numerical stability.
This regularizer penalizes degenerate solutions with low temporal variance, effectively promoting motion diversity. To prevent it from dominating the optimization once the model escapes the collapsed regime, the loss is truncated after convergence to a sufficiently diverse temporal distribution.

\begin{algorithm}[b]
\scriptsize 
\caption{Training Procedure}
\label{alg:training}
\SetAlgoLined
\DontPrintSemicolon
\SetKwInOut{Input}{Input}\SetKwInOut{Output}{Output}
\SetKwFunction{FwdDiff}{ForwardDiffusion}
\SetKwFunction{DenoiseLoss}{DenoisingLoss}
\SetInd{0.3em}{0.6em} 
\SetNlSty{}{\color{gray}\scriptsize}{\hspace{0.5em}}

\LinesNotNumbered
\Input{Pretrained teacher model $s_{\text{real}}$; Video data $\mathcal{D} = \{\mathbf{c},\mathbf{y}\}$; \\
Few-step denoising timesteps $\mathcal{T} = \{0, t_1, t_2, \dots, t_Q\}$}
\Output{Few-step generator $G$}
\LinesNumbered
\setcounter{AlgoLine}{0}
$G, s_{\text{fake}} \leftarrow \text{init}(s_{\text{real}})$; $\mathcal{\bar L}_{\text{reg}} \leftarrow 0$\;
\While{not converged}{
    \If{is generator update step}{
        \tcc{Update Generator G}
        $\mathbf{x} \gets G(\mathbf{z} \sim \mathcal{N}(0, I), \mathbf{c} \sim \mathcal{D})$\;
        $\mathcal{L}_{\text{KL}} \gets \text{DMDLoss}(s_{\text{real}}, s_{\text{fake}}, \mathbf{x})$ \tcp*{Eq.\eqref{eq:dmd}}
        $\mathcal{L}_{\text{temp}} \gets \text{TemporalLoss}(\mathbf{x})$ \tcp*{Eq.\eqref{eq:temp}}
        $\mathbf{y}_t \gets \FwdDiff((\mathbf{c}_y,\mathbf{y}) \sim \mathcal{D}, t \sim \mathcal{T})$\;
        $\mathbf{\hat y} \gets G(\mathbf{y}_t,\mathbf{c}_y )$\;
        $\mathcal{L}_{\text{reg}} \gets \text{RegressionLoss}(\mathbf{\hat y}, \mathbf{y})$ \tcp*{Eq.\eqref{eq:regloss}}
        
        $\mathbf{\omega}_{t,s} \gets f(\mathcal{\bar L}_{\text{reg}}, \mathcal{L}_{\text{reg}},t)$ \tcp*{Eq.\eqref{eq:weight}}
        $\mathcal{L}_G \gets \mathcal{L}_{\text{KL}} + \omega_{\text{temp}}\mathcal{L}_{\text{temp}} + \omega_{\text{reg}}\omega_{t,s}\mathcal{L}_{\text{reg}}$\;
        $G \leftarrow G - \eta_G \nabla_G \mathcal{L}_G$\;
        $\mathcal{\bar L}_{\text{reg}} \leftarrow (1-\alpha)\mathcal{\bar L}_{\text{reg}} + \alpha\mathcal{L}_{\text{reg}}$ \tcp*{EMA update}
    } 

    \tcc{Update online model $s_{\text{fake}}$}
    $\mathbf{x}_{t'} \gets \FwdDiff(\text{stop\_grad}(\mathbf{x}), t' \sim \mathcal{U}(0, 1))$\;
    $\mathcal{L}_{\text{denoise}} \gets \text{DenoisingLoss}(s_{\text{fake}}(\mathbf{x}_{t'}, t'), \text{stop\_grad}(\mathbf{x}))$\tcp*{Eq.\eqref{eq:mse_loss}}
    $s_{\text{fake}} \leftarrow s_{\text{fake}} - \eta_{s} \nabla_{s_{\text{fake}}} \mathcal{L}_{\text{denoise}}$\;
}
\end{algorithm}

\newcolumntype{C}[1]{>{\centering\arraybackslash}m{#1}}
\begin{table*}[t]
 \centering
 \caption{Quantitative comparison on two benchmarks (VBench2 and VBench1).
 Our method achieves the best overall score across all metrics and datasets. DMD$^*$ is denoted as our baseline method. In all result tables, bold text indicates the optimal result, while underlined text represents the suboptimal result.}
 \label{tab:merged_two_benchmarks}
 \setlength{\tabcolsep}{3pt}
 \resizebox{\textwidth}{!}{

  \renewcommand{\arraystretch}{1.25}

 \begin{tabular}{
   p{1.4cm} 
   C{1.0cm} 
   C{1.8cm} 
   C{1.2cm} 
   C{1.0cm} 
   |C{1.2cm}C{1.4cm}C{1.3cm}C{1.2cm}C{1.2cm}|C{1.2cm}
   |C{1.2cm}C{1.3cm}|C{1.2cm}
  }
  \toprule
  \multirow{3}{*}{Method} &
  \multirow{3}{*}{Steps} &
  \multirow{3}{*}{Resolution} &
  \multirow{3}{*}{Params} &
  \multirow{3}{*}{Time} &
  \multicolumn{6}{c|}{VBench2 Evaluation Dimension} &
  \multicolumn{3}{c}{VBench1 Evaluation Dimension} \\
  \cmidrule(lr){6-11} \cmidrule(lr){12-14}
  & & & & &
  Creati- vity & Common- sense & Controll- ability & Human Fidelity & Physics & \textbf{Total} &
  Quality Score & Semantic Score & \textbf{Total} \\
  \midrule
  Teacher & 50$\times$2 & 832$\times$480$\times$81 &  1.3B & 270s & 48.01 & 58.51 & 21.82 & 80.65 & 45.97 & 50.99 & 84.11 & 64.20 & 80.13 \\
  DMD$^*$\cite{yin2024one} & 4 & 832$\times$480$\times$81 &  1.3B & 10.8s & \underline{55.90} & 56.51 & 25.29 & \underline{86.75} & 43.72 & 53.63 & \textbf{85.55} & 61.10 & \underline{80.66} \\
  LCM\cite{luo2023latent} & 6 & 832$\times$480$\times$81 &  1.3B & 16.2s & 27.66 & 45.53 & 10.95 & 76.07 & 40.40 & 40.12 & 77.62 & 50.15 & 72.12 \\
  PCM\cite{wang2024phased} & 6 & 832$\times$480$\times$81 &  1.3B & 16.2s & 44.54 & 50.42 & 17.02 & 81.48 & 43.53 & 47.40 & 79.36 & 51.40 & 73.70 \\
  DCM\cite{lv2025dcm} & 6 & 832$\times$480$\times$81 &  1.3B & 16.2s & \textbf{61.31} & 52.71 & 16.93 & 83.99 & 44.07 & 51.80 & 79.27 & 52.16 & 73.92 \\
  rCM\cite{zheng2025large} & 4 & 832$\times$480$\times$81 &  1.3B & 10.8s & 46.40 & \textbf{60.79} & \underline{29.19} & 82.98 & \textbf{50.81} & \underline{54.03} & 84.80 & \underline{61.55} & 80.15 \\
  Ours & 4 & 832$\times$480$\times$81 &  1.3B & \textbf{7.8s} & 48.19 & \underline{59.07} & \textbf{29.39} & \textbf{88.26} & \underline{50.48} & \textbf{55.08} & \underline{85.27} & \textbf{65.65} & \textbf{81.35} \\
\bottomrule
Teacher & 50$\times$2 & 832$\times$480$\times$81 &  14B & 730s & 50.63 & 57.39 & 25.62 & 87.33 & 39.72 & 52.14 & 84.44 & 70.47 & 81.64 \\
    DMD$^*$ & 4 & 832$\times$480$\times$81 &  14B & 28s & 54.92 & 59.31 & \textbf{33.86} & 83.78 & 52.48 & 56.87 & 82.22 & 69.27 & 79.63 \\
    LCM & 6 & 832$\times$480$\times$81 &  14B & 42s & 36.71 & 51.28 & 19.87 & 71.55 & 43.04 & 44.49 & 80.33 & 53.84 & 75.03 \\
    DCM & 6 & 832$\times$480$\times$81 &  14B & 42s & \textbf{57.20} & 56.82 & 18.91 & 84.48 & 44.82 & 52.44 & 80.80 & 55.51 & 75.73 \\
  Ours & 4 & 832$\times$480$\times$81 &  14B & \textbf{22.2s} & 52.89 & \textbf{65.09} & 30.17 & \textbf{89.00} & \textbf{58.13} & \textbf{59.06} & \textbf{84.70} & \textbf{74.06} & \textbf{82.57} \\
  \bottomrule
 \end{tabular}
 }
 \label{tab:two_benchmarks}
 \vspace{-0.2in}
\end{table*}

\subsection{Frame-interpolation Inference}
\label{subsec:3.4}
We propose a frame interpolation strategy to accelerate video diffusion model inference, motivated by the operational dichotomy in denoising: high-noise steps capture coarse semantics, while low-noise steps refine details~\cite{teng2025magi,lv2025dcm}. Considering the correlation between the high-noise denoising stage and the inter-frame similarity of clean samples (visualized in Figure~\ref{fig:similarity}), our approach reduces the frame rate by half during the high-noise stage (e.g., first 2 of 4 steps). A lightweight, pre-trained UNet module then interpolates the sequence back to the full frame rate within the VAE's latent space. This allows the subsequent low-noise denoising to erase interpolation-induced artifacts. The UNet is pre-trained on real data via a regression loss to predict the features of an intermediate frame from its neighbors (appendix Sec.~\ref{sec:unet}). This method substantially cuts computational costs by shortening the effective sequence length for inference, with a negligible impact on perceptual quality.

\subsection{Overall Training Process}

The overall training pipeline is summarized in Algorithm~\ref{alg:training}. Both the student generator $G$ and the online model $s_\text{fake}$ used for computing the distribution-matching loss are initialized with the teacher model’s weights at the beginning of training. We follow the framework of DMD2~\cite{yin2024improved}. For the distribution matching loss, we only use fake videos generated by $G$ with text conditions. The online model $s_\text{fake}$ is updated using the two timescale update rule from DMD2. This means that before each update to the student model, we first optimize $s_\text{fake}$ for multiple steps using videos sampled from the student. This helps $s_\text{fake}$ adapt to the student's current output distribution. Our temporal regularization loss is calculated on the same videos generated for distribution matching. Therefore, this step adds no extra forward pass cost. The regression loss, however, is calculated using the real video dataset. As shown in Eq.~\ref{eq:ema}, we maintain a separate exponential moving average (EMA) mean for each denoising step of the student model to compute the adaptive weights. Overall, our training process adds only one additional forward pass per update to the student model. The final loss function is formulated as:
\begin{equation}
\vspace{-0.1in}
\mathcal{L}_G = \mathcal{L}_{\text{KL}} + \omega_{\text{reg}}\omega_{t,s}\mathcal{L}_{\text{reg}} + \omega_{\text{temp}}\mathcal{L}_{\text{temp}}.
\label{eq:generator loss}
\end{equation}
\vspace{-0.2in}

\section{Experiments}
\label{Sec:Experiments}

\begin{figure*}[t]
  \centering
  \includegraphics[width=\linewidth]{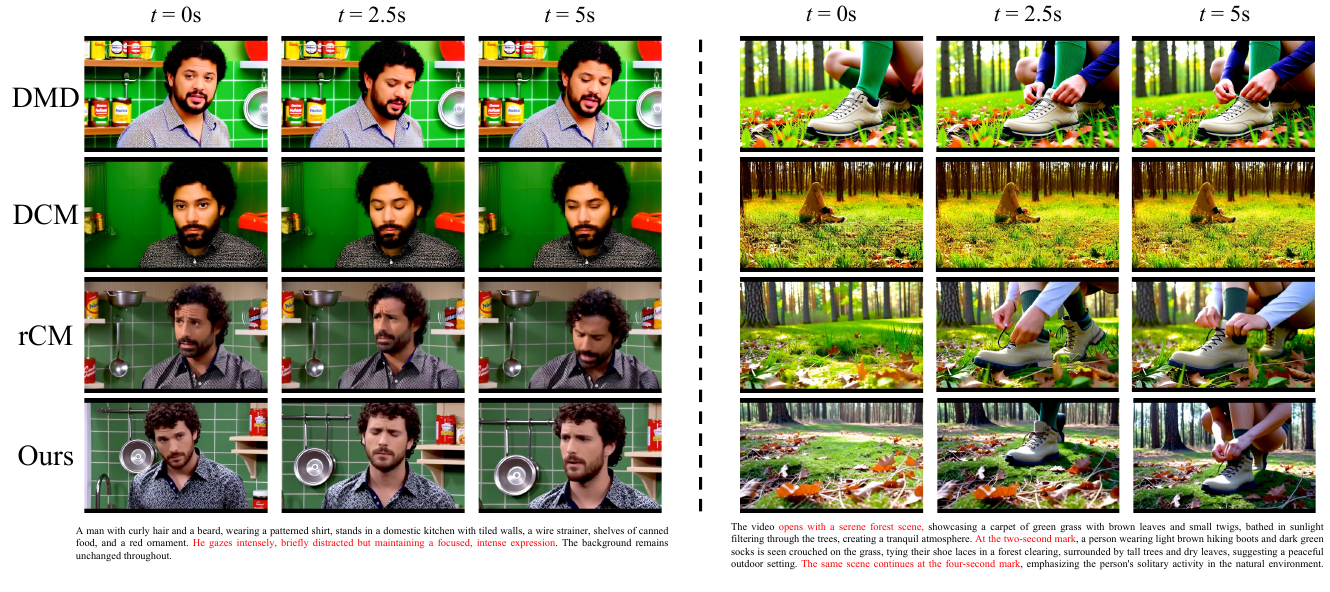}
  \vspace{-0.3in}
  \caption{Qualitative comparison. Baseline methods tend to produce oversaturated colors (left case) and exhibit stiff or static motion (right case), while our method generates videos with more natural color tones, smoother temporal dynamics, and scene transitions that better align with the prompt. The highlighted text in the prompt indicates key visual changes described in the scene. Zoom in for better visualization.}
  \label{fig:comparisons}
  \vspace{-0.15in}
\end{figure*}

\subsection{Experimental Setups}

We employ Wan2.1-T2V-1.3B and Wan2.1-T2V-14B as the teacher models for our experiments. Starting from their officially released pretrained weights, we distill them into a few-step denoising student model capable of generating 5-second, 16 fps, 832 × 480 resolution high-quality videos.

For training data, the distribution matching loss is conditioned on text annotations from a mixed dataset of open-source and proprietary videos, without direct use of the video data itself, following the methodology of DMD2~\cite{yin2024improved}. The regression loss, in contrast, is computed on a high-quality subset of 150,000 video samples, which we curated and cleaned from online sources. Further details on our data pre-processing are available in the appendix Sec.~\ref{sec:data}.

We use the AdamW optimizer with $lr$ = 2 × $10^{-6}$, an EMA decay factor $\alpha$ = 0.95, and $k$ = 3.0 for adaptive weight computation in Eq.~\ref{eq:temp}. The weights $\omega_{\text{reg}}$ and $\omega_{\text{temp}}$ in Eq.~\ref{eq:generator loss} are set to 2.0 and 0.05, respectively. The temporal regularization loss is truncated once it converges to around 0.6, and the teacher model employs a classifier-free guidance (CFG) scale of 5.0. The hyperparameter settings are discussed in detail in appendix Sec.~\ref{sec:hyperparameter}. All experiments are conducted on 24 GPUs.
\subsection{Evaluation Results}
\begin{figure}
    \centering
    \includegraphics[width=1\linewidth]{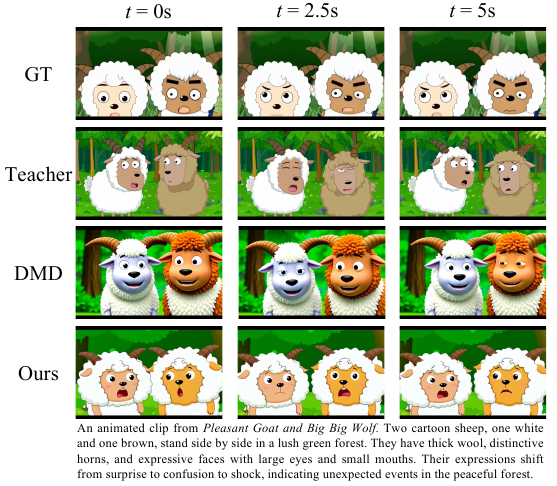}
    \vspace{-0.2in}
    \caption{The figure demonstrates that the adaptive-regression-loss enables effective distribution transfer during distillation. "GT" denotes ground-truth frames; both the Teacher and DMD models fail to generate specific anime-style videos consistent with the target domain.}
    \label{fig:cartoon}
    \vspace{-0.2in}
\end{figure}

\paragraph{Benchmark Evaluation.}
We conduct comprehensive quantitative evaluations on both VBench2~\cite{zheng2025vbench2} and VBench1~\cite{huang2023vbench} benchmarks to compare our approach against several state-of-the-art diffusion distillation methods\footnote{For methods with open-source weights, we use their released checkpoints. For others, we retrain the models on the same datasets following the original hyperparameter settings. All models are evaluated in a unified codebase with BF16 precision to ensure fair and consistent comparisons.}, including DMD (baseline). The results are summarized in Table~\ref{tab:two_benchmarks}. All methods are distilled from the same teacher model, using its officially released pretrained weights, and evaluated with the inference steps consistent with their respective training configurations.
Our method achieves state-of-the-art performance on both benchmarks in terms of the Total Score. Specifically, on VBench2, our approach shows substantial improvements in Human Fidelity compared to existing methods, demonstrating more precise motion alignment and perceptual realism. On VBench1, our model attains the highest Semantic Quality score, further confirming its superior generalization and generation fidelity across diverse prompts. Furthermore, we obtained the highest total scores on both models and benchmarks.
\vspace{-0.1in}
\paragraph{Qualitative Comparisons.}
We further evaluate the generalization capability of our distilled model by assessing its performance on high-quality prompts unseen during training. As illustrated in Figure~\ref{fig:comparisons}, our method demonstrates a significant reduction in color oversaturation while producing richer fine-grained details and smoother, more extensive motion. These improvements markedly enhance the perceptual realism of videos generated from complex, detail-rich prompts, offering a distinct advantage in human visual perception.
Furthermore, we investigate the adaptability of our method for downstream tasks through supervised fine-tuning, facilitated by the adaptive regression loss during distillation. As shown in Figure~\ref{fig:cartoon}, by fine-tuning on a small-scale animation dataset, our method achieves remarkable style transfer. The student model becomes capable of generating videos in a specific animation style that the original teacher model fails to produce. This result strongly validates the practical utility and adaptability of our approach.

\begin{figure}
    \centering
    \includegraphics[width=1\linewidth]{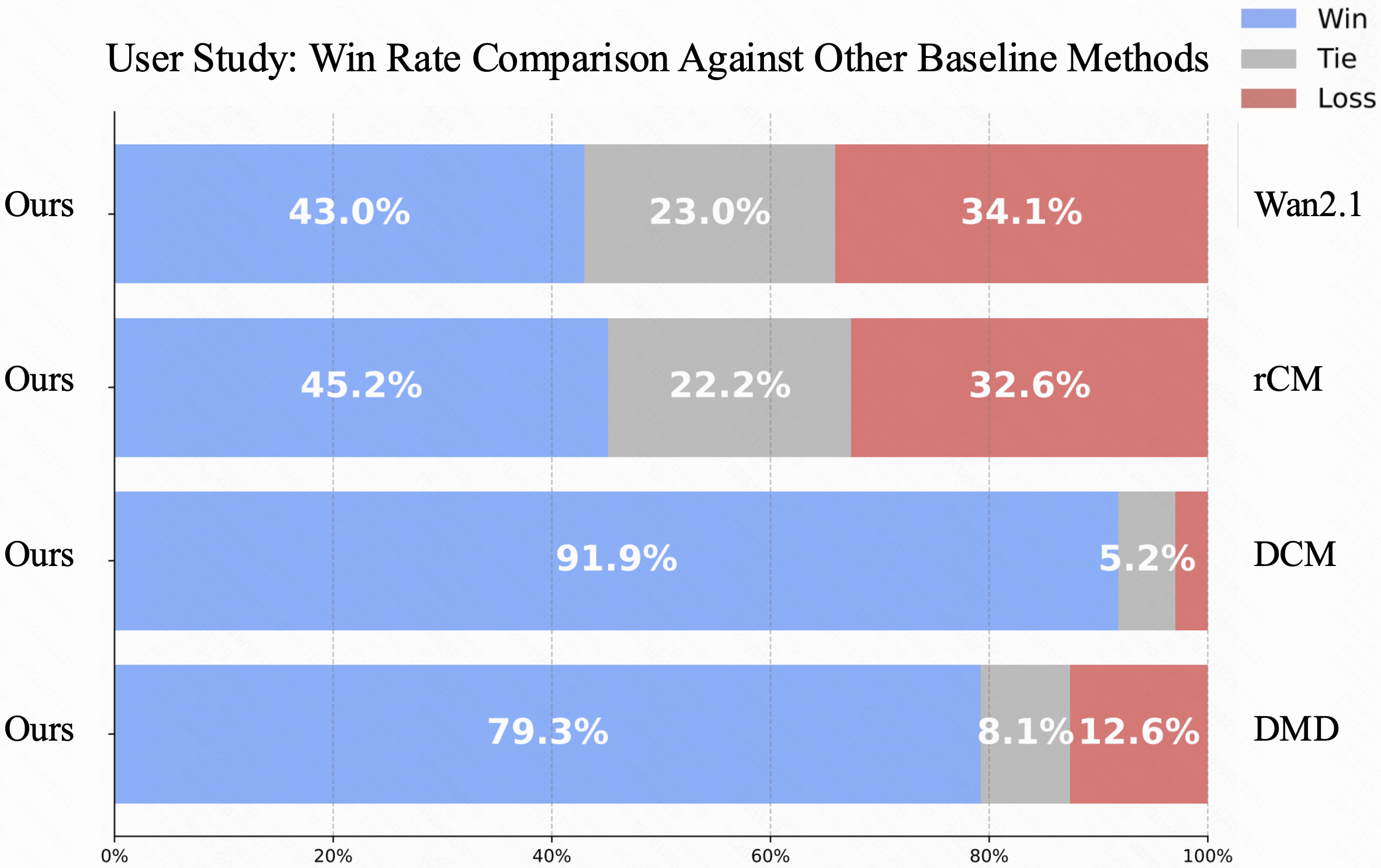}
    \caption{The figure presents a user study comparing our distilled student model against its teacher (Wan2.1) and other existing video diffusion models. The results demonstrate that our model achieves superior video quality, even surpassing its teacher model in terms of user preference, while operating at a significantly lower inference cost.}
    \label{fig:user study}
    \vspace{-0.25in}
\end{figure}

\vspace{-0.1in}
\paragraph{User Study.}
We conducted a human preference study to evaluate generation quality. Long-form prompts were created using a large language model, and from these prompts we produced a total of 180 comparison video samples. We recruited 12 professional, independent annotators; each paired sample was evaluated by at least three distinct annotators. The annotators were instructed to select the better video based on two criteria: visual quality and semantic alignment with the input prompt, with an option to rate them as comparable. For each comparison, we aggregated judgments by majority vote to produce the final preference. As shown in Figure~\ref{fig:user study}, our model consistently outperforms all baseline methods and is even preferred by users over its teacher model. This strongly demonstrates that our model achieves a more user-favored generation quality while operating at a much faster inference speed.
\subsection{Efficiency Analysis}
One single denoising step of the 1.3B model takes 2.7 seconds on one GPU, while inference at half frame rate reduces this time to 1.1 seconds. To further enhance efficiency, we employ a lightweight UNet-based interpolation network to upsample VAE-encoded video sequences. Specifically, our 4-step student model performs the first two denoising steps at half frame rate, and before the third denoising step, the interpolation network restores the sequence to its original frame rate. This interpolation process is extremely fast, taking only a few hundred milliseconds.This inference design effectively balances video quality and computational efficiency, achieving a 30\% acceleration in overall inference speed without compromising perceptual fidelity.

\subsection{Ablation Studies}
To validate the effectiveness of the proposed mechanisms, we conduct a series of evaluations based on the 1.3B DMD baseline. Specifically, we compare models trained with (1) temporal regularization, (2) regression loss, and (3) adaptive regression loss.

\vspace{-0.1in}
\paragraph{Adaptive Regression Loss.}
The Instance Preservation metric measures the temporal consistency of objects within a video. A lower score indicates instability, such as object merging, splitting, sudden appearance, or disappearance. As shown in Table~\ref{tab:ablation}, directly incorporating the regression loss leads to a significant drop in the Instance Preservation score compared to the model without it, indicating that the student model tends to produce more artifacts, such as tearing and object distortion. The examples in Figure~\ref{fig:two dogs} also illustrate this point. When replaced with the adaptive regression loss, the score improves markedly, even surpassing the DMD baseline and approaching the performance of the teacher model. This improvement stems from the model’s ability to focus on learning in reliable data regions while mitigating hallucinations in poorly aligned samples.

\vspace{-0.15in}
\paragraph{Temporal Regularization.}The Dynamic Degree metric evaluates the overall motion dynamics by estimating the optical flow magnitude across frames, where the score reflects the proportion of videos exhibiting meaningful motion among all test samples. As shown in Table~\ref{tab:ablation}, without the temporal regularization loss, DMD-based distillation suffers from a severe degradation in Dynamic Degree (over 10 percentage points). Introducing temporal regularization effectively restores motion diversity, ensuring that nearly all generated videos exhibit meaningful dynamics. Furthermore, when combined with the regression loss, the model maintains robustness and preserves a high level of motion fidelity.

\begin{table}[t]
  \centering
  \caption{Ablation study on the effectiveness of the proposed Adaptive Regression Loss and Temporal Regularization.
DMD serves as the baseline method, while \textbf{TR} and \textbf{AdaLoss} denote models equipped with the Temporal Regularization and Adaptive Regression Loss, respectively. \textbf{Full+VIF} refers to the model incorporating both components and frame-interpolation.}
  \label{tab:ablation}
  \begin{minipage}{0.5\textwidth}
    \centering
    \footnotesize 
    \setlength{\tabcolsep}{3pt} 
    \renewcommand{\arraystretch}{1.08} 
    \begin{tabular}{@{} 
        >{\raggedright\arraybackslash}m{1.1cm} 
        >{\centering\arraybackslash}m{1.0cm}
        >{\centering\arraybackslash}m{1.3cm}
        >{\centering\arraybackslash}m{1.0cm}
        | >{\centering\arraybackslash}p{1.25cm}
          >{\centering\arraybackslash}p{1.25cm} @{}}
      \toprule
      
  \multirow{4}{*}{Method} &
  \multirow{4}{*}{Steps} &
  \multirow{4}{*}{Resolution} &
  \multirow{4}{*}{Time} &
      \multicolumn{2}{c}{Evaluation Score} \\
      \cmidrule(lr){5-6}
      & & & & \shortstack{Instance\\Preservation} & \shortstack{Dynamic\\Degree} \\
      \midrule
      Teacher & $50\times2$ & $832\times480$ & 270s & \textbf{92.39} & 85.56 \\
      DMD & 4 & $832\times480$ & 10.8s & 88.88 & 72.22 \\
      +TR & 4 & $832\times480$ & 10.8s & 85.38 & \textbf{100.00} \\
      +TR+RegLoss & 4 & $832\times480$ & 10.8s & 83.04 & 78.61 \\
      +TR+AdaLoss & 4 & $832\times480$ & 10.8s & \textbf{92.39} & \underline{99.72} \\
      Full+VIF & 4 & $832\times480$ & \textbf{7.8s} & \underline{91.81} & 97.77 \\
      \bottomrule
    \end{tabular}
  \end{minipage}
  \vspace{-0.2in}
\end{table}

\vspace{-0.1in}
\section{Conclusion}
\vspace{-0.05in}
In this work, we addressed the key challenges of oversaturation and temporal inconsistency in the distillation of video diffusion models. We propose a novel framework that incorporates an adaptive regression loss to prevent spatial artifacts and correct color bias, alongside a temporal regularization loss to preserve motion dynamics and mitigate mode collapse. Combined with an inference-time acceleration strategy, our method enables stable, few-step (4-step), and fast video synthesis. Extensive experiments demonstrate that our approach significantly outperforms existing baselines on two VBench benchmarks, achieving superior perceptual fidelity and motion realism.

{
    \small
    \bibliographystyle{ieeenat_fullname}
    \bibliography{main}
}
\addtocontents{toc}{\protect\setcounter{tocdepth}{2}}
\clearpage
\setcounter{page}{1}
\maketitlesupplementary

\tableofcontents
\clearpage

\onecolumn
\begin{figure}[t]
  \centering
  \includegraphics[width=\textwidth]{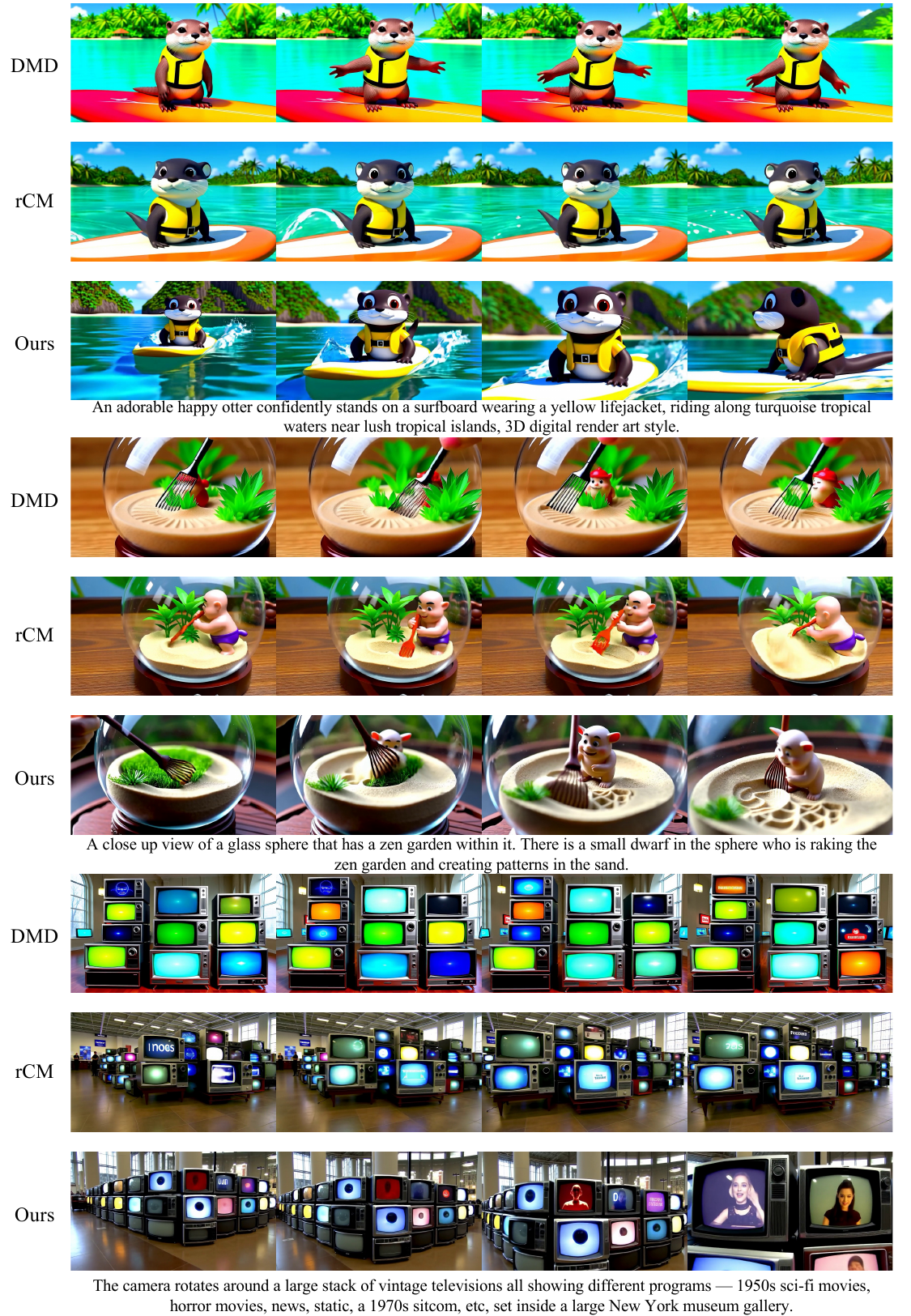}
  \caption{...}
  \label{fig:moviegen1}
\end{figure}

\clearpage

\begin{figure}[t]
  \centering
  \includegraphics[width=\textwidth]{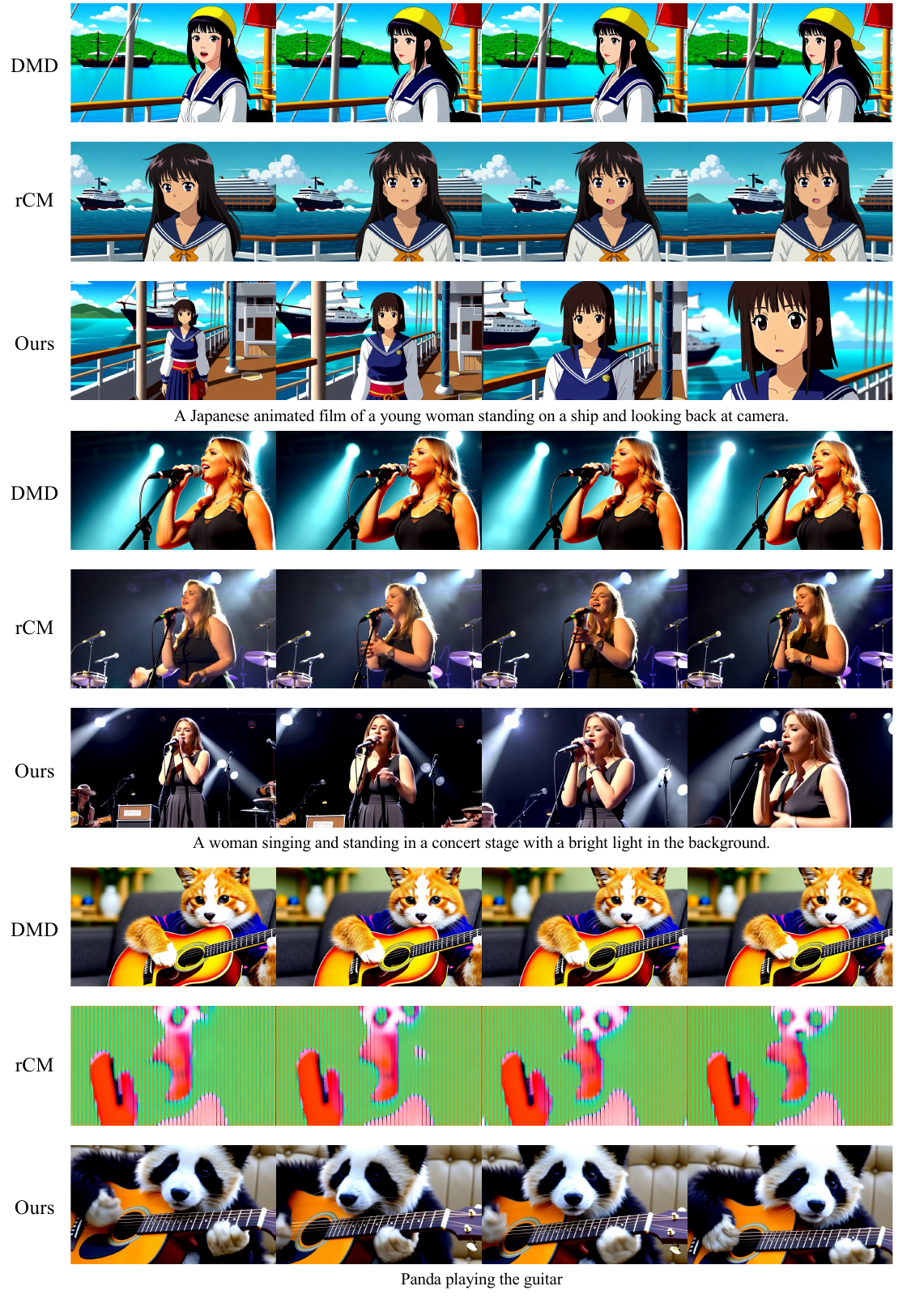}
  \caption{...}
  \label{fig:moviegen2}
\end{figure}

\clearpage

\begin{figure}[t]
  \centering
  \includegraphics[width=\textwidth]{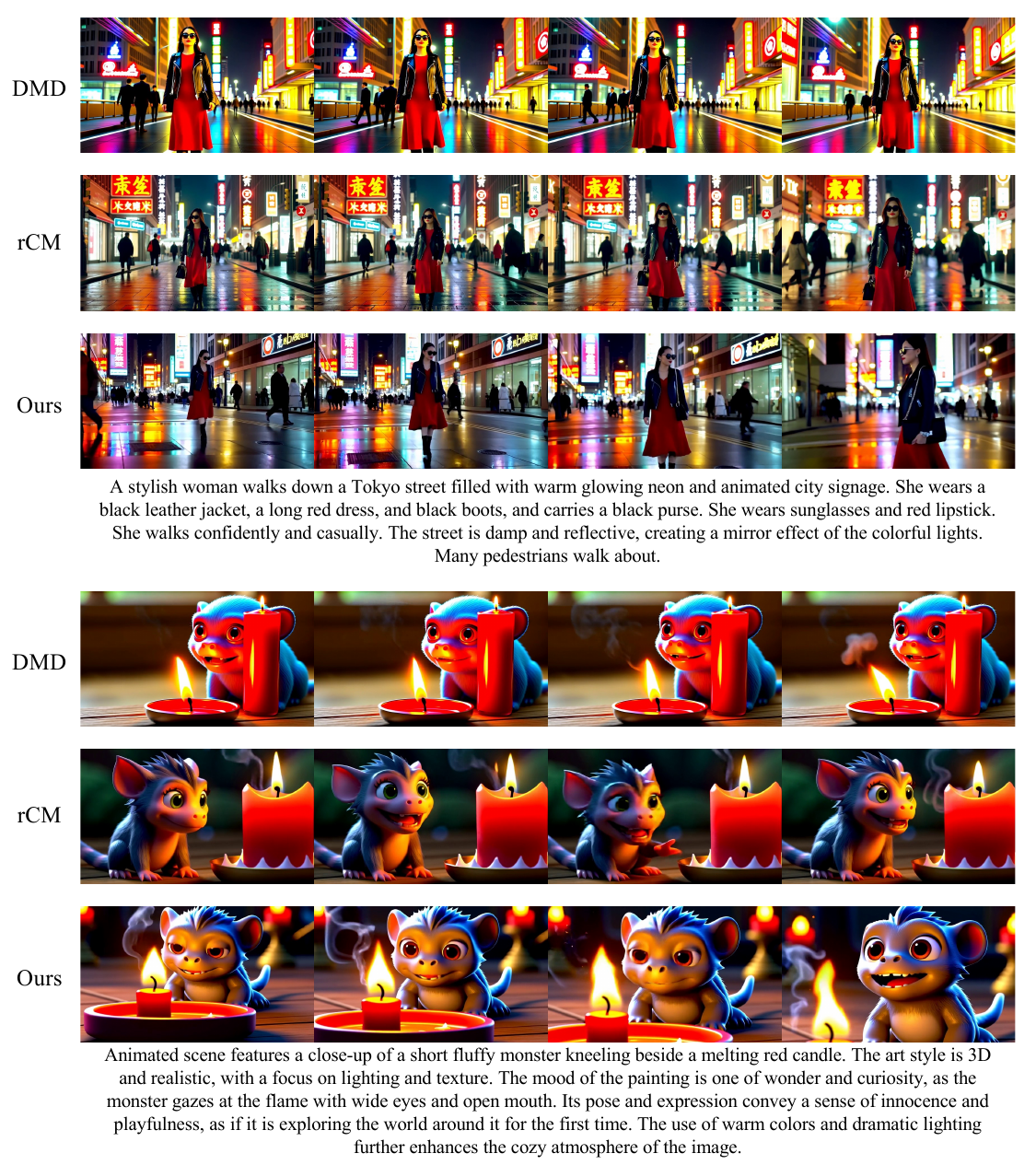}
  \caption{\textbf{Comparison with baselines on MovieGen prompts.} While DMD and rCM (a strong performer from Sec.~\ref{Sec:Experiments}) produce videos with low motion dynamics and oversaturated colors, our method resolves both issues. It achieves superior motion, well-calibrated colors, and excellent detail and stability.More video examples are included in the supplementary zip file. (Note: We have confirmed that any visual artifacts in some videos are not code-level issues.)}
  \label{fig:moviegen3}
\end{figure}

\clearpage
\twocolumn

\section{Training Data}
\label{sec:data}
This section details the pipeline for the collection, filtering, and annotation of the video dataset used in our training.
\subsection{Data Collection}
For the construction of a high-fidelity training dataset, we curated video data from two primary channels: (1) publicly available, open-source videos from the internet, and (2) our internal, proprietary archives. To achieve broad diversity and coverage, we deliberately included videos from a multitude of categories. These range from everyday scenes like lifestyle, travel, and food, to more specialized domains such as aerial cinematography, medical imaging, and professional sports, among many others.

\subsection{Data Filtering}
Our data filtering pipeline consists of two main stages designed to ensure the quality of the training videos.
\paragraph{Preliminary Filtering by Resolution.}
We first decode each video to determine its spatial resolution. To ensure sufficient source quality, we filter out all videos with a width or height below 720p (1280$\times$720). Subsequently, all remaining videos are uniformly resized to 480p for training, a process that helps preserve essential structural details from the high-resolution source material.
\paragraph{Frame-based Quality Assessment.}
From each remaining video, we uniformly sample four frames at fixed temporal positions (specifically, the 1st, 21st, 41st, and 61st frames). Each sampled frame must pass two complementary quality checks:
\begin{itemize}
    \item \textbf{Monochromatic Scene Detection:} We convert the frame to the HSV color space and compute the entropy of its hue histogram. If the hue entropy is below a predefined threshold of 0.60, the frame is classified as part of a monochromatic scene (e.g., all-black, all-white, or solid-color backgrounds), and the corresponding video is rejected.
    \item \textbf{Blur Detection:} We assess the sharpness of the frame by calculating the variance of its Laplacian operator. If the Laplacian variance is lower than 20.0, the frame is deemed excessively blurry (potentially due to out-of-focus or motion blur issues), and the video is eliminated.
\end{itemize}

\paragraph{Optical Flow Analysis for Motion Dynamics.}
To ensure that our training videos contain sufficient dynamic content, we perform an optical flow analysis on all videos that passed the initial screening. We use an optical flow algorithm to extract motion vector fields between consecutive frames and then compute the average motion magnitude. Videos with a magnitude below 0.2 are considered to have insignificant motion (e.g., static scenes or overly stable footage) and are excluded from the training set.

\paragraph{Temporal Consistency Filtering.}
To further enforce temporal coherence within video clips, we introduce an assessment based on frame-to-frame consistency. For the remaining videos, we calculate a temporal consistency score following the methodology of VBench1~[YourVBenchCitation]. We retain only the top 50\% of videos ranked by this score.
\begin{figure}
  \centering
  \includegraphics[width=1\linewidth]{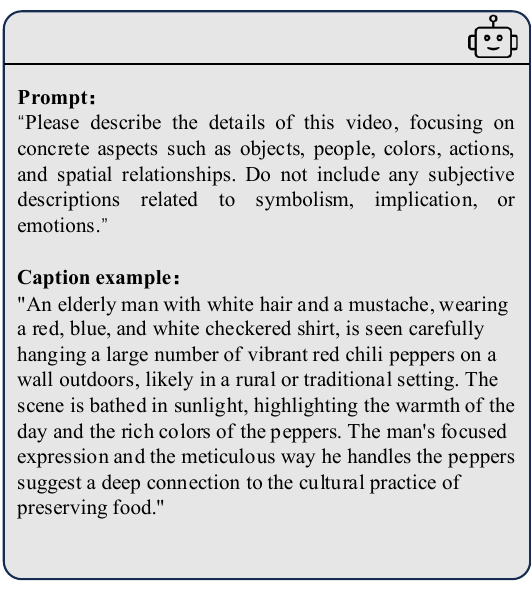}

    \caption{
    Caption model prompt and caption style examples.
    }
    \label{fig:caption} 
\end{figure}
\paragraph{Aesthetic Quality Filtering.}
Finally, we apply an aesthetic quality filter to ensure visual appeal. We employ a pre-trained visual aesthetic assessment model to score keyframes. Specifically, we uniformly sample 8 frames from each video, calculate their average aesthetic score, and keep only the top 50\% of videos based on this metric.

\subsection{Data Captioning}
For the filtered data, we employ a proprietary 7B-parameter captioning model for annotation. The prompt and style examples used for the annotation process are shown in Figure~\ref{fig:caption}.

\vspace{-0.1in}
\section{Quality Visualize}
\label{sec:Visualize}

To further evaluate our method's performance in practical video generation scenarios, we conducted a qualitative comparison on the official MovieGen prompt set against the DMD baseline and rCM, the latter of which performed favorably in our main experiments (Sec.~\ref{Sec:Experiments}). 
All methods generated videos under identical prompts and consistent sampling configurations to ensure a fair comparison. 
The results, as illustrated in Figures~\ref{fig:moviegen1}, \ref{fig:moviegen2}, and~\ref{fig:moviegen3}, demonstrate that our method consistently achieves superior and more stable performance across multiple key dimensions. 
Specifically, in terms of \textbf{motion dynamics}, our approach generates videos with more extensive action, cinematic camera movements, and greater fluidity compared to the baselines. 
Regarding \textbf{color saturation and overall style}, our method produces visuals with well-calibrated saturation, avoiding the oversaturation artifacts common in other approaches and resulting in a style that aligns more closely with natural video distributions. 
Furthermore, concerning \textbf{detail quality}, our model excels at preserving fine-grained textures and sharp structural contours, effectively mitigating issues such as blurring and local structural degradation. 
\begin{figure}
  \centering
  \includegraphics[width=1\linewidth]{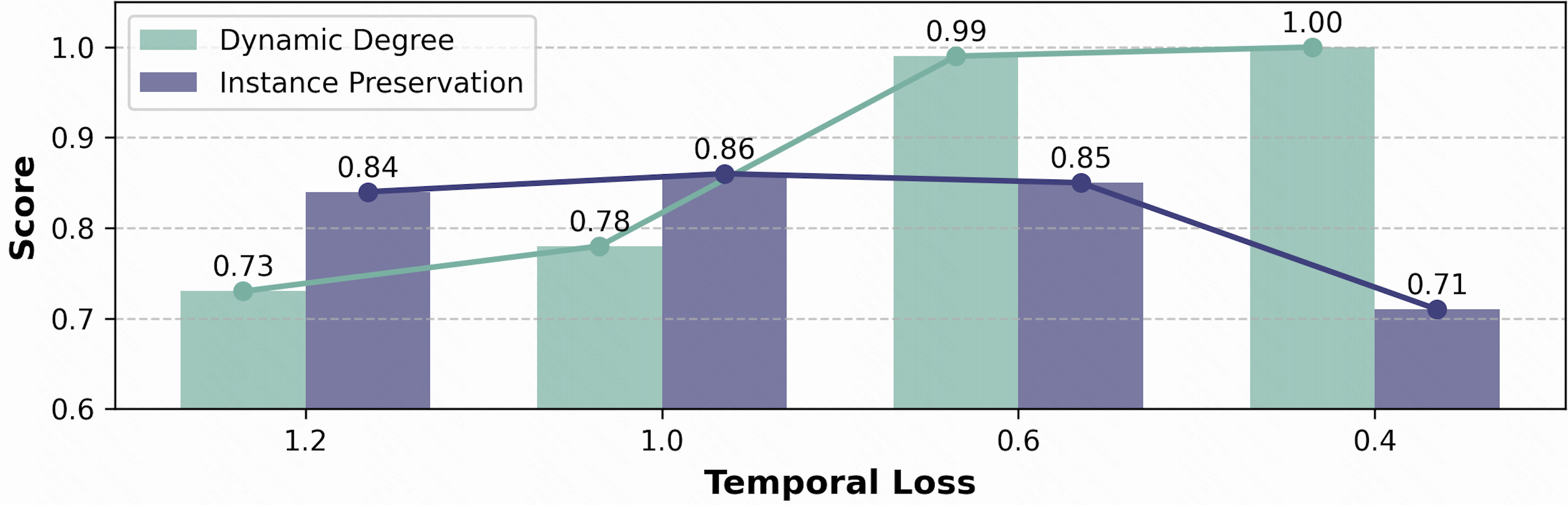}

    \caption{
        This figure illustrates the impact of the truncation threshold of the temporal regularization loss on model performance. The horizontal axis represents the truncation threshold of the regularization loss, while the vertical axis shows the motion score and the instance preservation score of the videos generated by the student model, respectively.
    }
    \label{fig:dynamic} 
    
\end{figure}

\begin{figure}
  \centering
  \includegraphics[width=1\linewidth]{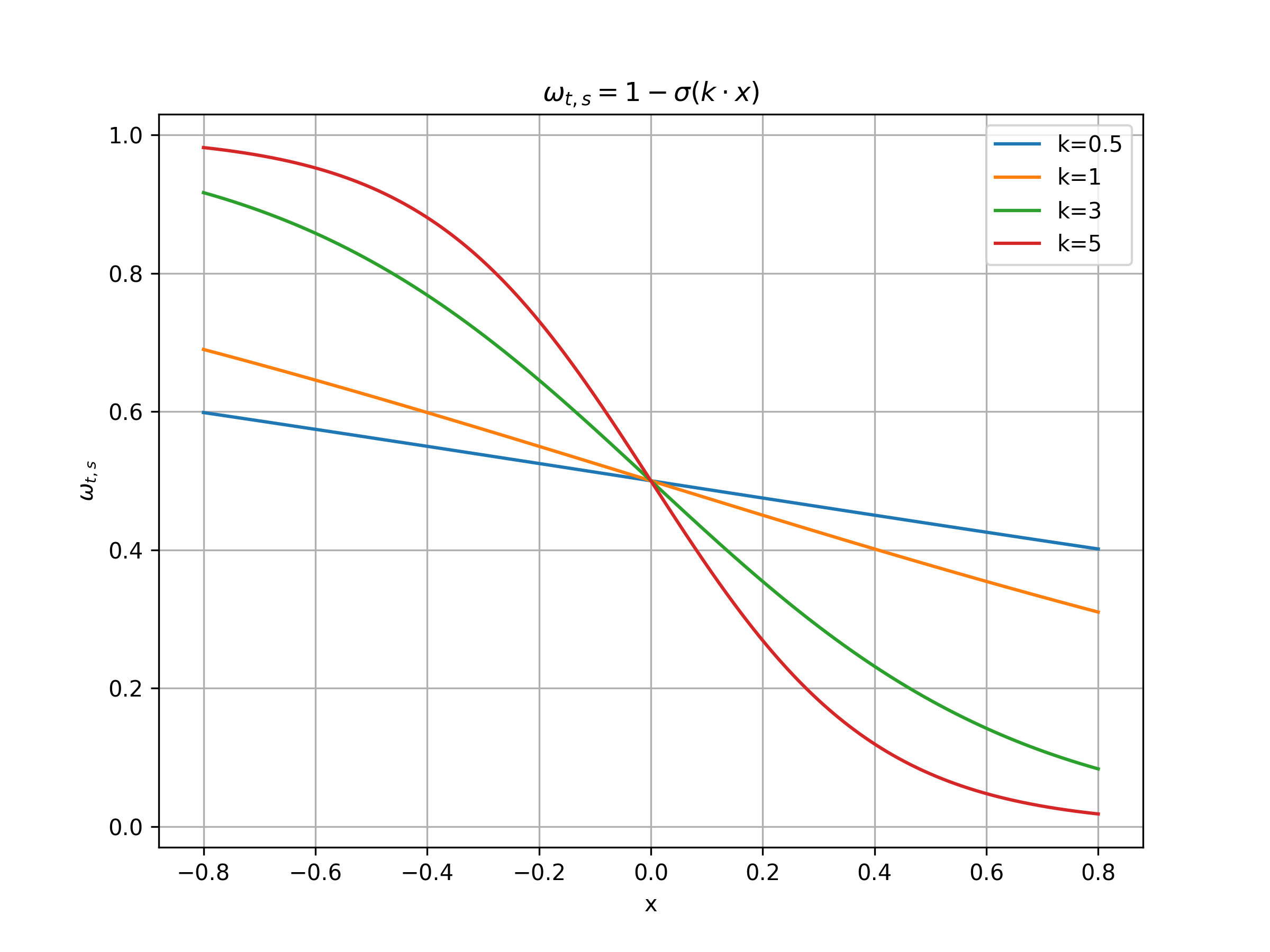}

    \caption{
        Visualization of the adaptive weight function in Eq.~\eqref{eq:weight} for different values of $k$. 
        The x-axis represents the deviation $\mathcal{L}_s - \bar{\mathcal{L}}_{s}$, while the y-axis represents the resulting adaptive weight.
    }
    \label{fig:k_visualization} 
    
\end{figure}
\begin{figure*}[t]
  \centering
  \includegraphics[width=\linewidth]{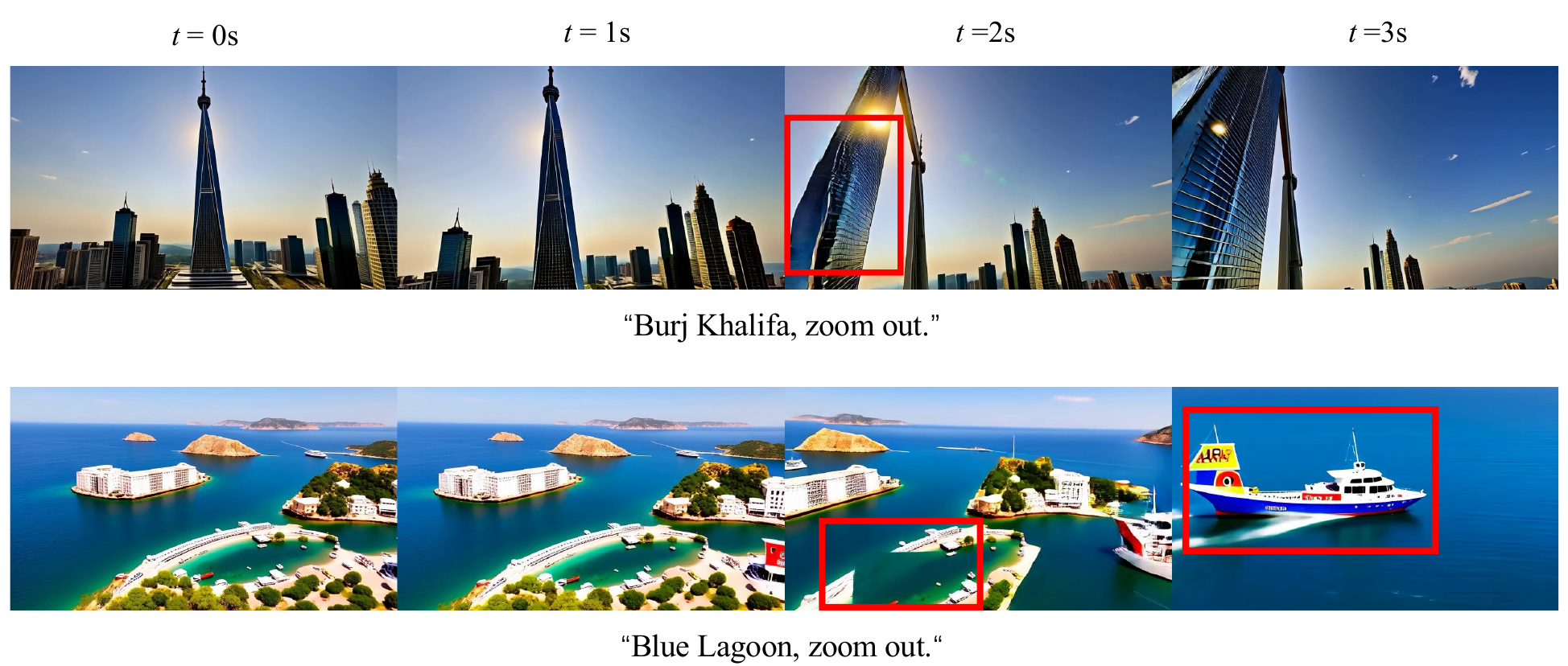}

    \caption{
        \textbf{Failure cases resulting from unclipped temporal regularization.} Without clipping, the student generator produces severe artifacts late in training. 
        \textbf{(Top)} After the first second, a drastic content shift occurs, accompanied by a noticeable distortion of the building (highlighted by the red box).
        \textbf{(Bottom)} The scene content abruptly vanishes at the two-second mark and is replaced by another major content shift at the three-second mark (highlighted by the red box).
        These phenomena, inconsistent with plausible camera motion, are clear manifestations of hallucinations. This highlights the necessity of clipping the temporal loss to prevent it from excessively amplifying inter-frame variance.
    }
    \label{fig:badcase} 
    
\end{figure*}

\begin{figure}[t]
    \centering
    \includegraphics[width=1\linewidth]{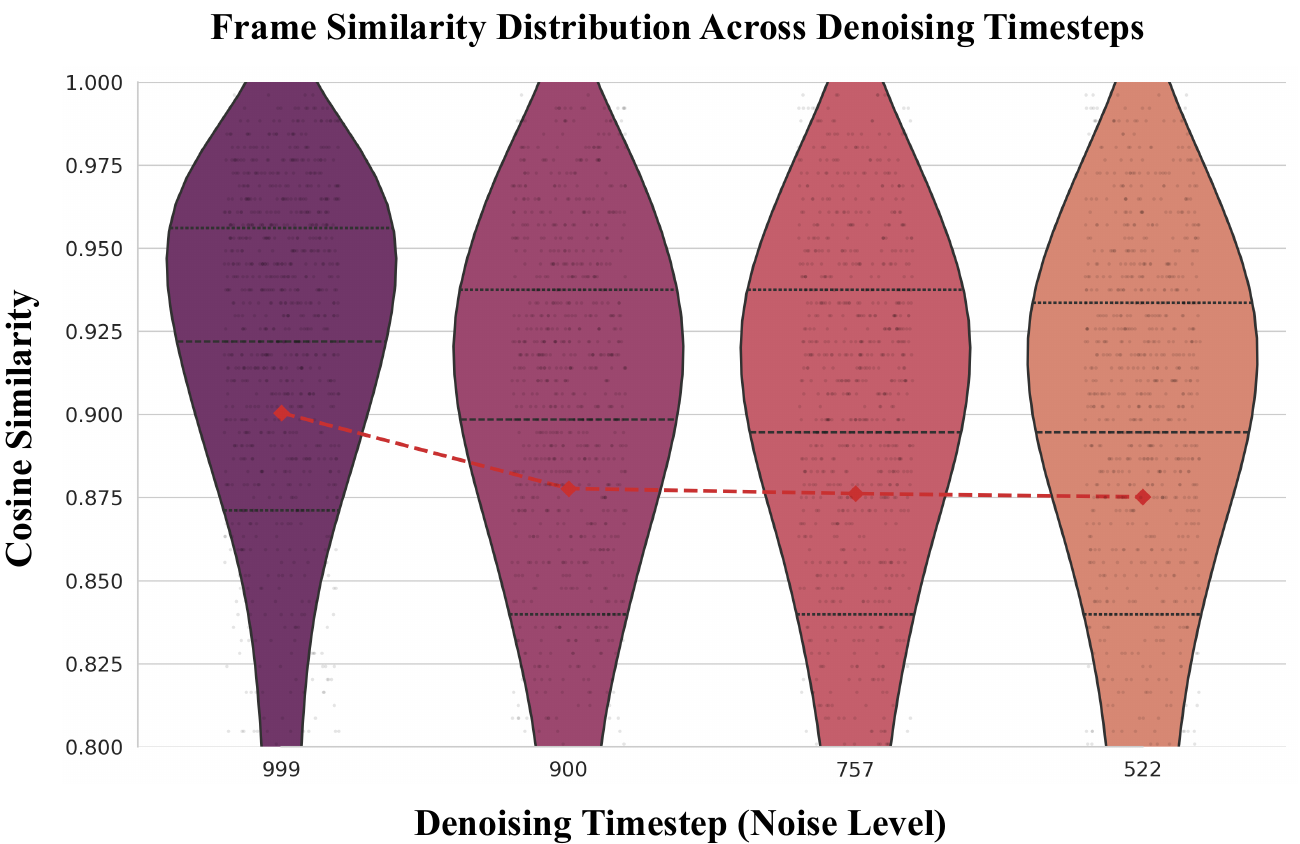}
    \caption{Violin plot showing the distribution of adjacent-frame cosine similarity across different denoising stages. The overall similarity is notably higher during the high-noise stage, which motivates our approach of halving the inference frame rate in this phase to reduce computational cost.    }
    \label{fig:similarity}
    \vspace{-0.15in}
\end{figure}
\section{Frame-interpolation Module}
\label{sec:unet}

\paragraph{Model Architecture.}
Our interpolation module is built upon a U-Net architecture designed to generate temporal intermediate frames between adjacent frames in the video's latent space. 
It takes as input the concatenated VAE-encoded latent features of two consecutive frames.

The network consists of a three-level downsampling encoder that progressively extracts high-level features, followed by a symmetric upsampling decoder that restores spatial resolution. 
Skip connections are employed to fuse high-resolution details from the encoder with deep semantic information from the decoder. 
The final output is an interpolated latent feature map with the same number of channels as a single frame's latent representation.

Specifically, the encoder is composed of three \texttt{ConvBlock} layers, each progressively expanding the channel dimensions and followed by max-pooling for downsampling. 
The decoder utilizes transposed convolutions for upsampling, concatenates the feature maps from the corresponding encoder level, and then fuses the information using another \texttt{ConvBlock}. 
A final 1$\times$1 convolution is applied to produce the C-channel intermediate frame.

\paragraph{Training Details.}
We train the U-Net interpolation module on a dataset of 150,000 real-world video clips, each 5 seconds long, with a resolution of 480p. The training procedure is as follows:

For each video, we first encode it into the latent space using the pre-trained VAE. 
A standard video in our latent space consists of 21 frames (corresponding to 81 frames at 16 fps in the pixel space over 5 seconds). 
From this 21-frame latent sequence, we uniformly downsample it to 11 frames. 
The U-Net module then takes these 11 frames and performs interpolation to restore the sequence to its original length.
The network is optimized by computing a regression loss between the interpolated latent video and the original 21-frame ground-truth latent video.

We use the AdamW optimizer with a learning rate of $1 \times 10^{-5}$, $\beta_1=0.9$, and $\beta_2=0.999$. The model is trained for 10,000 iterations with a batch size of 32.

\paragraph{Inference Process.}
During inference, the interpolator is invoked along the temporal dimension. 
For each pair of adjacent frames in the input sequence, the original frames are preserved. 
The two frames are then concatenated and fed into the U-Net to generate one intermediate frame, which is inserted in order between them. 
This process expands the sequence length from $F$ to $2F-1$, effectively doubling the video's frame rate.

The overall architecture is simple yet highly effective, capable of generating smooth and structurally consistent temporal intermediates in the latent space. 
It is suitable for applications such as video frame-rate up-sampling, transition generation, and interpolation tasks within diffusion models.

\section{Hyperparameter Discuss}
\label{sec:hyperparameter}
In this section, we discuss the hyperparameter values used during the training process. For most settings, we follow the configuration from Wan2.1. The specific parameters are as follows:

\begin{itemize}
    \item \textbf{Teacher Model:} The number of training timesteps is set to 1000. The classifier-free guidance scale is 5.0, and the timestep-shift is also 5.0.
    
    \item \textbf{AdamW Optimizer:} The parameters are set as follows: learning rate (lr) = $2.0 \times 10^{-6}$, $\beta_1 = 0.9$, $\beta_2 = 0.999$, and the maximum gradient norm (\texttt{max\_grad\_norm}) is 10.0.
    
    \item \textbf{Two-Timescale Update Rule:} The update frequency is set to 5. This means that the student generator is updated once for every five updates of the online model.
\end{itemize}

\subsection{Temporal Regularization}
The temporal loss is updated using the formulation in Eq.~\eqref{eq:temp}; we simply set $\epsilon$ to 1$\times 10^{-6}$.
Theoretically, this loss function incentivizes a continuous increase in the temporal variance of the model's output, as a larger variance results in a smaller negative logarithm and thus a lower loss.
Consequently, the loss lacks a natural convergence mechanism.
This can lead to numerical instability, excessively large gradients, and potentially exploding gradients.
Furthermore, the model's output can be amplified indefinitely, causing anomalous temporal variations that manifest as excessive jitter or noise in the generated videos.
This regularization term may also conflict with the primary training objective; for instance, it creates an optimization paradox if the main task promotes smoothness while this term encourages variance.

Empirically, we observed that without any clipping, this loss causes the model to generate videos with severe frame jumps or hallucinatory artifacts in the later stages of training (as shown in Figure~\ref{fig:badcase}).

Therefore, it is necessary to clip this loss once it falls below a certain value.
To establish a reasonable clipping threshold, we first computed this loss on the VAE-encoded latents of 4,000 videos generated by the teacher model.
The average value was found to be approximately 1.5.
Based on this observation, we experimented with three distinct clipping thresholds: 1.2, 1.0, 0.6 and 0.4.
We trained the model for a sufficient and equal number of epochs under each setting.
The dynamic degree score and instance preservation score of the videos generated by the student are presented in Figure~\ref{fig:dynamic}.
We observed that setting the threshold to 0.6 significantly improves motion dynamics while maintaining object generation quality.
Furthermore, to ensure that the distribution matching and adaptive regression losses remain the dominant sources of gradients during distillation, we found that setting the weight $\omega_\text{temp}$ for this regularization term to 0.05 provides an effective balance.At this weight, the temporal loss typically converges to near the clipping threshold during training, while its weighted magnitude remains a small fraction compared to the other two losses.

\subsection{Adaptive Regression Loss}
We now discuss the selection of key hyperparameters.

First, the parameter $\alpha$ in Eq.~\eqref{eq:ema} governs the contribution of historical losses to the exponential moving average (EMA); we adopt a commonly used value of 0.95.

Next, in Eq.~\eqref{eq:weight}, the parameter $k$ controls the slope of the sigmoid function. This, in turn, determines how sensitively the adaptive weight responds to the deviation of the current loss $\mathcal{L}_s$ from its historical trend $\bar{\mathcal{L}}_{s}$.
Notably, for low-noise steps, the absolute value of the regression loss $\mathcal{L}_s$ is typically very small (often below 0.01). At this stage, the guidance from the regression loss is inherently limited, as the input already contains significant information from the real image.
Consequently, the output of the sigmoid function remains close to 0.5, making the weight largely insensitive to the value of $k$.
Therefore, our analysis primarily focuses on the impact of $k$ during the high-noise denoising stages.

As visualized in Figure~\ref{fig:k_visualization}, which plots the weight function for several values of $k$ (where the x-axis is $\mathcal{L}_s - \bar{\mathcal{L}}_{s}$), we identified two desired behaviors.
\textbf{First}, during the initial training phases, when the distribution gap is large, $\mathcal{L}_s - \bar{\mathcal{L}}_{s}$ can reach values between 0.6 and 0.8. We want to avoid excessively penalizing these data points (i.e., the weight should not be too close to 0 in this range).
\textbf{Second}, in the later stages, when $\mathcal{L}_s - \bar{\mathcal{L}}_{s}$ has largely converged to the [-0.3, 0.3] interval, the weight function should still retain sufficient discriminative power to differentiate between samples.
Through experimentation, we found that setting $k=3.0$ provides an effective trade-off that satisfies both conditions.

Finally, for $w_{\text{reg}}$ in Eq.~\eqref{eq:generator loss}, we set its value to 2.0. This choice normalizes the weight to approximately 1.0 for data points where the current loss is near the historical average (i.e., $\mathcal{L}_s - \bar{\mathcal{L}}_{s} \approx 0$).

\end{document}